# Deep Learning Approach for Very Similar Objects Recognition Application on Chihuahua and Muffin Problem

Enkhtogtokh Togootogtokh[1#], Amarzaya Amartuvshin[2*]
[#]Department of Computer Science, Mongolian University of Science and Technology
[*]Department of Mathematics, National University of Mongolia
[1]enkhtogtokh.java@gmail.com, [2]amarzaya@num.edu.mn

**Corresponding Author:**
　Name: Enkhtogtokh Togootogtokh
　Email: enkhtogtokh.java@gmail.com
　Mobile: +886-0970717184
　Address: Department of Computer Science, Mongolian University of Science and Technology (MUST), Ulaanbaatar, Mongolia

**Abstract**
　　We address the problem to tackle the very similar objects like "Chihuahua or muffin" problem to recognize at least in human vision level. Our regular deep structured machine learning still doesn't solve it. We saw many times for about year in our community the problem. Today we proposed the state-of-the-art solution for it. Our approach is quite tricky to get the very high accuracy. We propose the deep transfer learning method which could be tackled all this type of problems not limited to just "Chihuahua or muffin" problem. It is the best method to train with small data set not like require huge amount data.

*Keywords*—Deep Learning; Deep Transfer Learning; Recognition for Chihuahua and Muffin; Deep Learning for Very Similar Objects Recognition, Chihuahua or Muffin Problem

# 1. Introduction

We strongly inspired from these research works [1],[2],[3]. How can human easily recognize such visual recognition problems like it is muffin or Chihuahua ?. As far as we know our visual perception and abstract thought has been trained by 540 million years of data and 100 thousand years of data. And we have transferred them into our new inheritance. Answers for newborn baby mathematic skill, and color, visual objects recognition are explained using those facts. So far our machine learning is still not sure about it on such task. Our strong motivation from here to develop the such artificial model using deep learning technology.  For general recognition task, convolutional neural network (ConvNet) has the best performance. It brings us to research in general about two important things : deep transfer learning and ConvNet. And important question is how to optimally combine them as transferred convolutional neural network for the general problem not limited to solve only one case for Chihuahua or Muffin. Since if we just propose a method to solve only this case, these type of problems are never end. As example in Fig. 1, machine learning still does not recognize well about which is labradoodle or fried chicken.  Another important thing is they are special case in real life problems we mostly find general huge data about them like 100,000 thousand any dogs but not so easy to find that much huge data about these special cases. Other words, if our deep artificial architecture can have good performance on not huge data like 500 Chihuahuas and 500 muffins, that must be very important. How is our proposed method the best for it and how is our deep transferred ConvNet the most well optimized, you could easily see in Experimental Result section. In human vision, we might have a some visual pattern for any objects, which means as example we know dog's head shape pattern like two symmetry eyes, nose has some more distance from eyes, and so on. Like very well preserved geometry pattern might be in 3D space. Then how about our artificial model, it can give us these things. One very excited way is to use pre-trained information like we already trained the general visual pattern as human does. Then we naturally save the energy as once we face the one special problem, we don't do erase all the information and then train from scratch. Instead, we just increment the those special knowledge.  We use pre-trained information as ImageNet weights [4]. It has been trained on 1 million data with 1000 categories like preserving the possibly good pattern for common objects. Possibly good pattern means here at least trained from any angle and tried to cover all general cases for common objects.

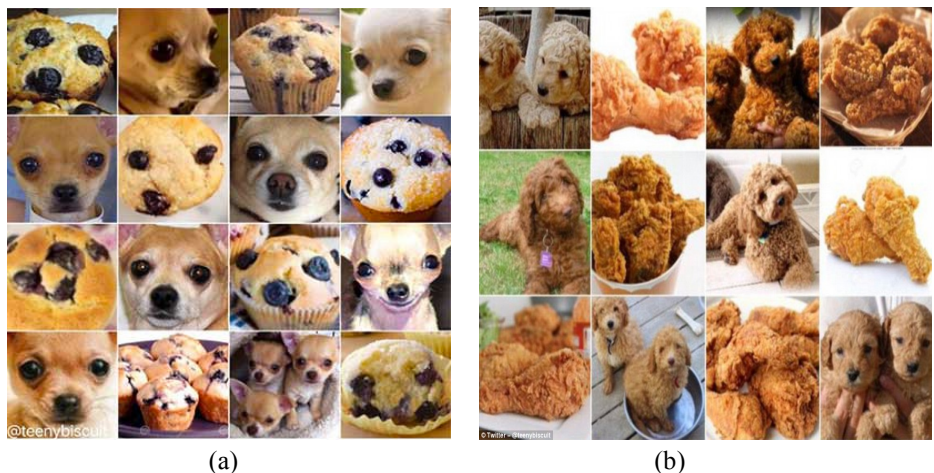

(a)          (b)
Figure 1. (a) Chihuahua and muffin, (b) Labradoodle and fried chicken

In this research, we tackle to recognize the very similar objects like human does. The key contributions of the proposed work are summarized as follows.
- It can work on small data.
- Not require as very big processing power.
- Optimized transferred deep learning technique.
- More than 97% accuracy to recognize such very similar objects.

The rest of the paper is organized as follows. The proposed method is explained in section 2. The experiment results and discussion is presented in section 3. Finally, Section 4 provides the conclusions and future work.

**2. The Proposed method**

In this section we discuss the deep transfer learning. First, we choose right pre-trained weights as ImageNet weights [4]. Then we propose the architecture. And we discuss the training part.

**2.1. Architecture**

We propose the 20 layers architecture like VGG19 [6] in Fig 2.    In detail it has following layers:
1. Convolutional Block-1: Two conv layers with 64 filters each. output shape:112 x 112 x 128
2. Conv Block-2: Two conv layers with 128 filters each. output shape: 56 x 56 x 256
3. Conv Block-3: Four conv layers with 256 filters each. output shape: 28 x 28 x 512
4. Conv Block-4: Four conv layers with 512 filters each. output shape: 14 x 14 x 512
5. Conv Block-5: Five conv layers with 512 filters each. output shape: 7 x 7 x 512
6. Fully connected layer 1: output shape: 1x 1 x 4096
7. Fully connected layer 2 : output shape: 1x 1 x 4096
8. Output(predictions): output shape: 1x1x1000 (For ImageNet)

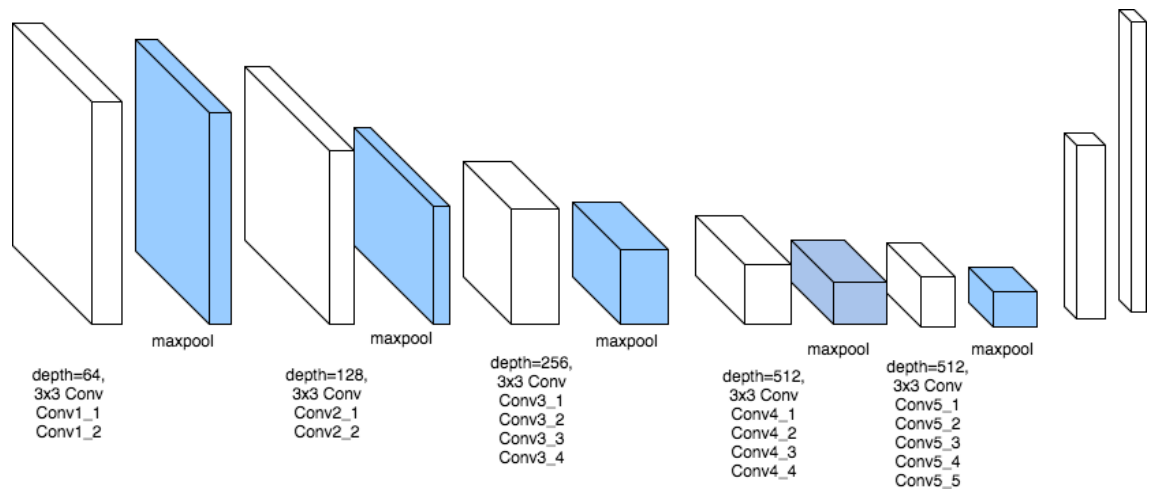

Figure. 2. The proposed architecture

We have already described the architecture above as we have created the graph for convolutional blocks of VGG19 like network loaded with imagenet pretrained weights. We only need to classify between Chihuahua and muffin, we shall take only the initial 5 convolutional blocks. Fine-tuning is one of the most common way to solve problems using AI and Deep learning. Pay attention here, for fine-tuning during this experiment, we say that we are happy with the weights of these layers so, don't change these weights, rather we shall change the weights of the small network that we will be added in the next step.

We use just 500 images of each chihuahua and muffin to train. In computer vision, data augmentation is the important to train from given data set and common ways are as flipping images, rotations, random crops to add small variations without damaging the central object. We should consider a mechanism to perform these kinds of data augmentations. We shall show how we are able to achieve more than 90% accuracy with little training data during pretraining. Another important thing to note here is we normalize by dividing each pixel value in all the images by 255, since we are using 8bit images so each pixel value is now between 0 and 1. This is also loosely called pre-processing of input images for VGG network. Our imagenet weights have also been obtained using the same normalization. Hence, if you miss this, you will get very bad predictions.

## 2.2. Fine-Tuning

As described in [2], we consider the fine-tuning technique as the weights of the pretrained network by continuing the backpropagation. Since that ConvNet earlier layers contain more generic features like edge detectors, color blobs, and so on and later layers have more specific details features of the special classes, there are two ways to fine-tune. First, we can do fine-tune all the layers, and second way is to keep some of the earlier layers fixed to consider overfitting and only fine-tune some higher-level layers of the architecture. In our case, second way is more comfortable because of differentiation of very similar objects features.

To choose right type of transfer learning, we have to consider two important factors as size of new dataset and its similarity to the original dataset. As described earlier in previous paragraph our architecture features are generic in early portions and specific in later portions, we listed common rules as described in [2] as 4 scenarios:

   a) *New dataset is small and similar to original dataset*. The data is small then it is not a good idea to fine-tune the ConvNet due to overfitting concerns. The data is similar to the original data, we expect higher-level features in the ConvNet to be relevant to this dataset as well. Hence, the best idea might be to train a linear classifier on the ConvNet.
   b) *New dataset is large and similar to the original dataset*. Since we have more data, we can have more confidence that we won't overfit if we were to try to fine-tune through the full network.
   c) *New dataset is small but very different from the original dataset*. Since the data is small, it is likely best to only train a linear classifier. Since the dataset is very different, it might not be best to train the classifier form the top of the network, which contains more dataset-specific features. Instead, it might work better to train the SVM classifier from activations somewhere earlier in the network.
   d) *New dataset is large and very different from the original dataset*. Since the dataset is very large, we may expect that we can afford to train a ConvNet from scratch. However, in practice it is very often still beneficial to initialize with weights from a pretrained model. In this case, we would have enough data and confidence to fine-tune through the entire network.

Fine – Tuning is consisted of 4 main steps such that data augmentation, setup the network, add the top as number of classes, and specify the full model input, output, and optimizer, loss functions as described in Algorithm 1.

---

Algorithm 1: *FineTuning* ($D_{tr}$, $D_{test}$)

INPUT: $D_{tr}$ – training data set, $D_{test}$ – testing data set, $W_{pre}$ - pretrained weights, $W_{pre}$, $D_{tr}$, $D_{test}$ ∈ ***M***
OUTPUT: $W_{ij}$ ∈ ***M***

// Image transformation to augment data
01.   $I_{tr}$ ∈ $D_{tr}$, $I_{test}$ ∈ $D_{test}$, $flip(I_{tr})$, $flip(I_{test})$ ∈ $D_{tr}$, $rotate(I_{tr})$, $rotate(I_{test})$ ∈ $D_{test}$

// Setup network
02.   model = VGG19(include_top=false, weights=none);
03.   model.add(Conv2D(5,5));

// Add the top as number of classes
04. z = model.output
05. z = Dense(128)(z)
06. z = GlobalAveragePooling2D()(z)
07. z = Dropout(0.3)(z)
08. predictions = Dense(2, activation='softmax')(z)

// Specify the complete model input and output, optimizer and loss
09. checkpoint = ModelCheckpoint($W_{ij}$, monitor='val_loss')
10. model_complete = Model(inputs=model.input, outputs=predictions)
11. model_complete.load_weights($W_{pr}$)
12. model_complete.compile(loss="categorical_crossentropy", optimizer=optimizers.SGD(lr=0.0001, momentum=0.9))
13. $n_{ti}$ = 500, $n_{vi}$ = 300, $b_{size}$=10 // Number of training image, number of validation image, batch size
14. $s_{epoch}$ =   $n_{ti}$ / $b_{size}$ // steps per epoch
15. $v_{step}$= $n_{vi}$ / $b_{size}$
16. model_complete.fit_generator(
        $D_{tr}$,
         steps_per_epoch= $v_{step}$,

    epochs=25,
    validation_data = $D_{test}$,
    validation_steps= $v_{step}$)
17. **return** $W_{ij}$

Note: M is matrix space

## 2.3. Train

First, we don't load Imagenet pretrained weights. In previous Section 2.2, we discuss about fine-tuning technique as how to implement it efficiently. Our training process consists of 2 steps as first pretrain, and then fine-tuning. Here we stated pretraining process as shown in Algorithm 2. In pretraining, there are 4 important parts such that data augmentation, network setup, add the top as number of classes, and specify the complete model input and output. In fine-tuning, we load the Imagenet weights saved in pretraining step and the network remain the same as described in Section 2.1. During fine-tuning, our model is efficient enough so we don't need to change the weights much. Here, Stochastic Gradient Descent (SGD) is the best option, however, there are other methods as Adam, Adagrad and so on. Finally, after all these processes, we achieve close to 98% with only 500 examples of each class which is very impressive.

---

Algorithm 2: *Pretrain* ($D_{tr}$, $D_{test}$)

INPUT:  $D_{tr}$ – training data set, $D_{test}$ – testing data set, $W_{pre}$- pretrained weights, $W_{pre}$, $D_{tr}$, $D_{test}$ ∈ **M**
OUTPUT: $W_{ij}$ ∈ **M**

// Image transformation to augment data
02.  $I_{tr}$ ∈ $D_{tr}$, $I_{test}$ ∈ $D_{test}$, $flip(I_{tr})$, $flip(I_{test})$ ∈ $D_{tr}$, $rotate(I_{tr})$, $rotate(I_{test})$ ∈ $D_{test}$

// Setup network
02.  model = VGG19(include_top=false, weights=none);
03.  model.add(Conv2D(5,5));

// Add the top as number of classes
04. z = model.output
05. z = Dense(128)(z)
06. z = GlobalAveragePooling2D()(z)
07. z = Dropout(0.3)(z)
08. predictions = Dense(2, activation='softmax')(z)

// Specify the complete model input and output, optimizer and loss
09. checkpoint = ModelCheckpoint($W_{ij}$, monitor='val_loss')
10. model_complete = Model(inputs=model.input, outputs=predictions)
11. model_complete.load_weights($W_{pr}$)
12. model_complete.compile(loss="categorical_crossentropy", optimizer=optimizers.Adagrad())
13. $n_{ti}$ = 500, $n_{vi}$ = 300, $b_{size}$=10 // Number of training image, number of validation image, batch size
14. $s_{epoch}$ =  $n_{ti}$ / $b_{size}$ // steps per epoch
15. $v_{step}$= $n_{vi}$ / $b_{size}$
16. model_complete.fit_generator(
   $D_{tr}$,
   steps_per_epoch= $v_{step}$,
   epochs=25,
   validation_data = $D_{test}$,
   validation_steps= $v_{step}$)
17. **return** $W_{ij}$

Note: M is matrix space

## 3. Experimental Results

In this section, the experimental results of the proposed method are presented and reviewed. Our experimental setup consists of Mac Book PRO, Core i7-2600 @3.6 GHz processor, and 16 GB Ram.

### A. Datasets

We have collected the 1000 images from the internet in 2 categories as muffin and Chihuahua including 200 images from Oxford pet animal dataset [7] as shown in Figure 3. Please note here, all resources as images from the internet is for research purpose only, we don't own any of them. ImageNet [5] also includes 1750 Chihuahua and 1335 various type of muffin images already.

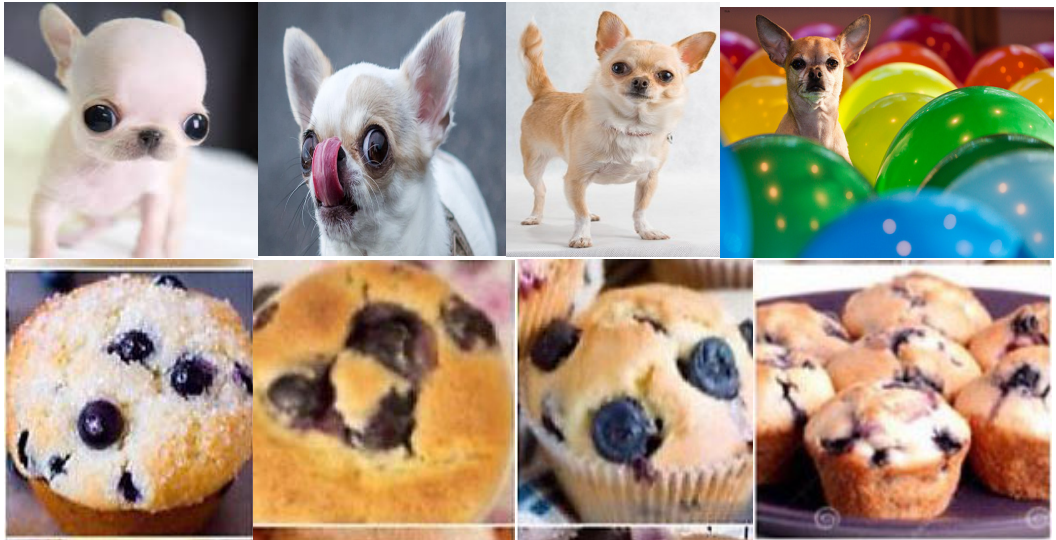

Figure. 3. Chihuahua and Muffin dataset

### B. Recognition Accuracy

We listed the some interesting most similar images with accuracy rate from our proposed method in simple Table 1.

Table 1. A some most similar cases of Chihuahua and Muffin accuracy results

| Objects | Accuracy | Error |
|---------|----------|-------|
| 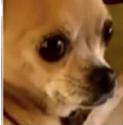 | 0.981 | 0.12 |
| 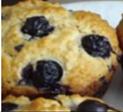 | 0.983 | 0.11 |
| 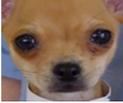 | 0.99 | 0.01 |
| 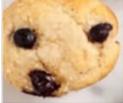 | 0.98 | 0.12 |
| 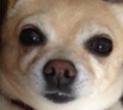 | 0.97 | 0.13 |
| 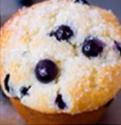 | 0.98 | 0.12 |
| 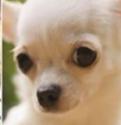 | 0.99 | 0.11 |
| 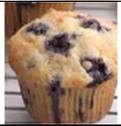 | 0.99 | 0.11 |
| | 0.98 | 0.12 |

In Table 2, we compare our results to other famous object recognition API's from Google, Amazon, IBM, Microsoft, Clarifai using cloudy vision [8]. Please note here, these other approaches mostly have not exact results as example in Chihuahua case, they recognize as in general dog, or cake. And another thing is those methods mostly don't even recognize the objects in Top-1 results. In Table 2, we collect the statistic data from their recognition results whatever they have in more close results. It is mostly in Top-5, Top-10 results.

Table 2. Our proposed method comparison results with others

| Objects | Amazon | Clarifai | Google | IBM | Microsoft | Our method |
|---|---|---|---|---|---|---|
| 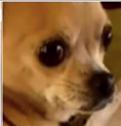 | 0.98 (Ch) | 1.0 (Dog) | 0.97 (Dog) | 0.77 (Dog) | 1.0 (Dog) | 0.98 (Ch) |
| 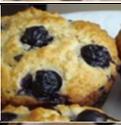 | 0.78 (B) | 0.97 (Cake) | 0.89 (Muffin) | 0.92 (Bread) | 0.33 (Bread) | 0.98 (Muffin) |
| 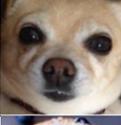 | 0.93 (Ch) | 1.0 (Dog) | 0.97 (Dog) | 0.79 (Dog) | 1.0 (Dog) | 0.97 (Ch) |
| 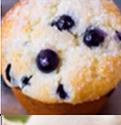 | 0.93 (Muffin) | 0.95 (Muffin) | 0.92 (Muffin) | 0.82 (Bread) | 0.74 (Bread) | 0.98 (Muffin) |
| 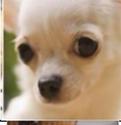 | 0.98 (Ch) | 1.0 (Dog) | 0.97 (Dog) | 0.96 (Dog) | 1.0 (Dog) | 0.99 (Ch) |
| 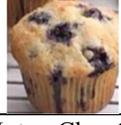 | 0.98 (Muffin) | 0.97 (Cake) | 0.92 (Muffin) | 0.99 (Bread) | 0.59 (Bread) | 0.99 (Muffin) |

Note : Ch – Chihuahua

## 4. Conclusion

The proposed method is fully solved the very similar object recognition like muffin or Chihuahua. It is the right solution for the such problem. Because first, human like solution is provided here as our knowledge is incrementing everyday especially when we face in difficult recognitions between very similar visual objects. And second, we already inherited base knowledge from our ancestors like in our case artificial pretrained knowledge. Third, it saves a lot of energy from machine since we only train our special case and increase the knowledge about the that special problem. It should never end. Since we learn everyday it does not matter how big knowledge or how a little we are gaining there.

## References

[1] George, Daniel, Hongyu Shen, and E. A. Huerta. "Deep Transfer Learning: A new deep learning glitch classification method for advanced LIGO." *arXiv preprint arXiv:1706.07446*(2017).
[2] Transfer Leaning. http://cs231n.github.io/transfer-learning/.
[3] Transfer Learning. http://cv-tricks.com/keras/fine-tuning-tensorflow/
[4] http://www.dailymail.co.uk/news/article-3487538/Is-Chihuahua-muffin-Hilarious-photos-dogs-look-like-food-sweeping-internet.html. March 2016.
[5] IMAGENET. http://www.image-net.org/
[6] Simonyan, Karen, and Andrew Zisserman. "Very deep convolutional networks for large-scale image recognition." *arXiv preprint arXiv:1409.1556* (2014).
[7] Oxford Pet Animal Dataset. http://www.robots.ox.ac.uk/~vgg/data/pets/
[8] Cloudy Vision. https://github.com/goberoi/cloudy_vision

## BIOGRAPHIES

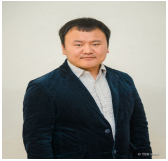
**Enkhtogtokh Togootogtokh** is Professor and Artificial Intelligence Researcher at the Department of Computer Science, Mongolian University of Science and Technology (MUST), Mongolia. He acquired his Masters and Bachelors respectively from National University of Mongolia, Mathematics and Computer Science School, Mongolia. He gained his Ph.D degree from National Central University, Taiwan. His research interests are in AI, deep learning, machine learning, image processing and computer vision, HCI.

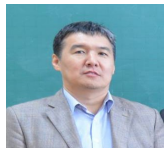
**Amarzaya Amartuvshin** is a Professor at the Department of Mathematics, School of Arts and Science, National University of Mongolia, Mongolia. He acquired his Masters and Bachelors respectively from National University of Mongolia, Mathematics and Computer Science School, Mongolia. He gained his Ph.D degree from Tokyo Metropolitan University, Japan. His research interests are in AI, Deep Learning, Differential Geometry and Manifold theory.